\documentclass{article}

\PassOptionsToPackage{numbers, compress}{natbib}

\usepackage[preprint]{neurips_2024}

\usepackage[utf8]{inputenc} 
\usepackage[T1]{fontenc}    
\usepackage{hyperref}       
\usepackage{url}            
\usepackage{booktabs}       
\usepackage{amsfonts}       
\usepackage{nicefrac}       
\usepackage{microtype}      

\usepackage{etoc}

\usepackage{amsmath}
\usepackage{amsthm}
\usepackage{amsfonts}
\usepackage{amssymb}
\usepackage{mathtools}
\usepackage[most]{tcolorbox}

\newcommand{\matr}[1]{\mathbf{#1}}

\usepackage[capitalize,noabbrev]{cleveref}

\hypersetup{
  colorlinks,
  citecolor=orange,
  linkcolor=red,
  urlcolor=blue
}

\theoremstyle{plain}

\theoremstyle{definition}

\theoremstyle{remark}

\DeclareMathOperator*{\argmin}{arg\,min}

\usepackage{wrapfig}

\usepackage{listings}

\definecolor{codecellbg}{RGB}{245,245,245}
\definecolor{codecellborder}{RGB}{231,231,231}
\definecolor{codered}{RGB}{171,49,42}
\definecolor{codegreen1}{RGB}{80,126,127}
\definecolor{codegreen2}{RGB}{55,126,33}
\definecolor{codeblue}{RGB}{0,0,245}
\definecolor{codeorange}{RGB}{255,128,0}
\definecolor{codepurple}{RGB}{156,47,246}

\lstdefinestyle{pythonStyle}{
    language=Python,
    basicstyle=\ttfamily,
    breaklines=true,
    showstringspaces=false,
    commentstyle=\color{codegreen1},
    keywordstyle=\color{codegreen2},
    stringstyle=\color{codered},
    numberstyle=\tiny\color{gray},
    escapeinside={(*@}{@*)},
    numbers=left,
    numbersep=5pt,
    morekeywords={self},
    breakatwhitespace=false,
    breaklines=true,
    keepspaces=true,
    showspaces=false,
    showstringspaces=false,
    showtabs=false,
    tabsize=4,
    postbreak=\mbox{\textcolor{red}{$\hookrightarrow$}\space},
}

\newtcblisting{jupyterCell}{
    colback=codecellbg,
    colframe=codecellborder,
    boxrule=0.5pt,
    arc=0mm,
    leftrule=3pt,
    rightrule=0.5pt,
    toprule=0.5pt,
    bottomrule=0.5pt,
    left=5mm,
    right=2mm,
    top=0mm,
    bottom=0mm,
    listing only,
    listing options={
        style=pythonStyle,
    },
}

\newtcbox{\codebox}[1][]{
    on line,
    colback=codecellbg,
    colframe=codecellborder,
    boxrule=0.5pt,
    arc=0pt,
    outer arc=0pt,
    left=2pt,
    right=2pt,
    top=2pt,
    bottom=2pt,
    boxsep=0pt,
    fontupper=\ttfamily,
    #1
}
\newcommand{\code}[1]{\codebox{#1}}

\tikzstyle{mynode}=[thick,draw=black,circle,minimum size=1]

\usepackage{geometry}
\usepackage{array}
\usepackage{amssymb}
\usepackage{booktabs}

\title{\textsc{jpc}: Flexible Inference for \\Predictive Coding Networks in JAX}

\author{%
  Francesco Innocenti \\
  School of Engineering and Informatics \\
  University of Sussex \\
  \texttt{F.Innocenti@sussex.ac.uk} \\
  \And
  Paul Kinghorn \\
  School of Engineering and Informatics \\
  University of Sussex \\
  \texttt{p.kinghorn@sussex.ac.uk} \\
  \AND
  Will Yun-Farmbrough \\
  School of Engineering and Informatics \\
  University of Sussex \\
  \texttt{wy77@sussex.ac.uk} \\
  \And
  Miguel De Llanza Varona \\
  School of Engineering and Informatics \\
  University of Sussex \\
  \texttt{m.de-llanza-varona@sussex.ac.uk} \\
  \AND
  Ryan Singh \\
  School of Engineering and Informatics \\
  University of Sussex \\
  \texttt{rs773@sussex.ac.uk} \\
  \And
  Christopher L. Buckley \\
  School of Engineering and Informatics \\
  University of Sussex \\
  \texttt{c.l.buckley@sussex.ac.uk} \\
}

\begin{document}

\maketitle

\begin{abstract}
  We introduce \textsc{JPC}, a \textbf{J}AX library for training neural networks with \textbf{P}redictive \textbf{C}oding. \textsc{JPC} provides a simple, fast and flexible interface to train a variety of PC networks (PCNs) including discriminative, generative and hybrid models. Unlike existing libraries, \textsc{JPC} leverages ordinary differential equation solvers to integrate the gradient flow inference dynamics of PCNs. We find that a second-order solver achieves significantly faster runtimes compared to standard Euler integration, with comparable performance on a range of datasets and network depths. \textsc{JPC} also provides some theoretical tools that can be used to study PCNs. We hope that \textsc{JPC} will facilitate future research of PC. The code is available at 
  \begin{center}
  \begin{tcolorbox}[colback=white, colframe=black, width=0.5\textwidth, height=0.6cm, valign=center, boxrule=0.1mm, sharp corners]
    \url{www.github.com/thebuckleylab/jpc}
  \end{tcolorbox}
  \end{center}
\end{abstract}

\section{Introduction} \label{intro}

In recent years, predictive coding (PC) has been explored as a biologically plausible learning algorithm that can train deep neural networks as an alternative to backpropagation \cite{van2024predictive, millidge2021predictive, millidge2022predictivereview, salvatori2023brain}. However, with a few recent notable exceptions \cite{legrand2024pyhgf, pinchetti2024benchmarking}, there has been a lack of unified open-source implementations of PC networks (PCNs) which would facilitate research and reproducibility\footnote{We also acknowledge earlier libraries such as \href{https://github.com/infer-actively/pypc}{\texttt{pypc}} and \href{https://github.com/RobertRosenbaum/Torch2PC}{\texttt{Torch2PC}} \cite{rosenbaum2022relationship}.}.

In this short paper, we introduce \textsc{JPC}, a \textbf{J}AX library for training neural networks with \textbf{PC}. \textsc{JPC} provides a simple, fast and flexible API for training a variety of PCNs including discriminative, generative and hybrid models. Like JAX, \textsc{JPC} follows a fully functional paradigm close to the mathematics, and the core library is <1000 lines of code. Unlike existing implementations, \textsc{JPC} leverages ordinary differential equation solvers (ODE) to integrate the gradient flow inference dynamics of PCNs. \textsc{JPC} also provides some theoretical tools that can be used to study and potentially identify problems with PCNs.

The rest of the paper is structured as follows. After a brief review of PC (\S\ref{pc}), we showcase some empirical results showing that  a second-order ODE solver can achieve significantly faster runtimes than standard Euler integration of the gradient flow PC inference dynamics, with comparable performance on different datasets and networks (\S\ref{runtime}). We then explain the library's core implementation (\S\ref{implement}), before concluding with possible extensions (\S\ref{conclusion}).

\begin{tcolorbox}[colback=teal!5!white,colframe=teal!75!black, title=\phantom{blabla}]
    \section{Predictive coding: A primer} \label{pc}
    Here we include a minimal presentation of PC necessary to get started with \textsc{JPC}. The reader is referred to \cite{van2024predictive, millidge2021predictive, millidge2022predictivereview, salvatori2023brain} for reviews and to \cite{buckley2017free} for a more formal treatment. \\
    
    PCNs are typically defined by an energy function which is a sum of squared prediction errors across layers, and which for a standard feedforward network takes the form
    \begin{equation}
        \mathcal{F} = \sum_{\ell=1}^L ||\mathbf{z}_\ell - f_\ell(\matr{W}_\ell \mathbf{z}_{\ell-1})||^2
        \label{eq1}
    \end{equation}
    where $\mathbf{z}_\ell$ is the activity of a given layer and $f_\ell$ is some activation function. We ignore multiple data points and biases for simplicity. \\
    
    To train a PCN, the last layer is clamped to some data, $\mathbf{z}_L \coloneq \mathbf{y}$. This could be a label for classification or an image for generation, and these two settings are typically referred to as \textit{discriminative} and \textit{generative} PC. The first layer can also be fixed to some data serving as a ``prior'', $\mathbf{z}_0 \coloneq \mathbf{x}$, such as an image in a supervised task. In unsupervised training, this layer is left free to vary like any other hidden layer. \\
    
    The energy (Eq. \ref{eq1}) is then minimised in a bi-level fashion, first w.r.t. the activities (inference) and then w.r.t. the weights (learning)
    \tikzset{
        freenode/.style={
            circle,
            draw=red!50,    
            fill=red!10,  
            thick
        },
        fixednode/.style={
            circle,
            draw=black,    
            fill=white,    
            thick
        }
    }
    
    \vspace{0.2cm}
    \begin{minipage}{0.005\textwidth}
        \begin{tikzpicture}[x=0.6cm,y=0.6cm]
          \foreach \N [count=\lay,remember={\N as \Nprev (initially 0);}]
                       in {2,3,3,2}{ 
            \foreach \i [evaluate={
              \x=\N/2-\i; 
              \y=-\lay; 
              \prev=int(\lay-1);
              \opacity=0.3+0.8*rand}]
                         in {1,...,\N}{ 
              \ifnum \lay=1
                  \ifnum \i=2 
                      \node[fixednode] (N\lay-\i) at (\x,\y) {};
                      \node[anchor=east] at (N\lay-\i.west) {\hspace{1mm}$\mathbf{x}$};
                  \else
                      \node[fixednode] (N\lay-\i) at (\x,\y) {};
                  \fi
              \else
                  \ifnum \lay=4
                      \ifnum \i=2 
                          \node[fixednode] (N\lay-\i) at (\x,\y) {};
                          \node[anchor=east] at (N\lay-\i.west) {\hspace{1mm}$\mathbf{y}$};
                      \else
                          \node[fixednode] (N\lay-\i) at (\x,\y) {};
                      \fi
                  \else
                      \node[freenode, opacity=\opacity] (N\lay-\i) at (\x,\y) {};
                  \fi
              \fi
              \ifnum\Nprev>0
                \foreach \j in {1,...,\Nprev}
                  \draw[->] (N\prev-\j) -- (N\lay-\i);
              \fi
            }
          }
        \end{tikzpicture}
    \end{minipage}
    \hspace{0.75cm}
    \begin{minipage}{0.4\textwidth}
        \begin{equation}
            \begin{aligned}
                \textit{Infer:  }
                \argmin_{\textcolor{red!50}{\textstyle\mathbf{z}_\ell}} \mathcal{F}
            \end{aligned}
        \label{eq2}
        \end{equation}
    \end{minipage}
    \hspace{0.2cm}
    \begin{minipage}{0.005\textwidth}
        \begin{tikzpicture}[x=0.6cm,y=0.6cm]
          \foreach \N [count=\lay,remember={\N as \Nprev (initially 0);}]
                       in {2,3,3,2}{ 
            \foreach \i [evaluate={\x=\N/2-\i; \y=-\lay; \prev=int(\lay-1);}]
                         in {1,...,\N}{ 
              \node[fixednode] (N\lay-\i) at (\x,\y) {};
              \ifnum\Nprev>0 
                \foreach \j in {1,...,\Nprev}{ 
                  \pgfmathsetmacro{\randomseed}{int(100*rand)}
                  \pgfmathsetseed{\randomseed}
                  \pgfmathsetmacro{\thickness}{0.6 + 0.3*rand}
                  \draw[->, blue!50, line width=\thickness pt] (N\prev-\j) -- (N\lay-\i);
                }
              \fi
            }
          }
        \end{tikzpicture}
    \end{minipage}
    \hspace{0.8cm}
    \begin{minipage}{0.4\textwidth}
        \begin{equation}
            \begin{aligned}
                \textit{Learn:  }
                \argmin_{\textcolor{blue!50}{\textstyle\matr{W}_\ell}} \mathcal{F}
            \end{aligned}
        \label{eq3}
        \end{equation}
    \end{minipage}
    \vspace{0.2cm}

    The inference dynamics are generally first run to convergence until $\Delta \mathbf{z}_\ell \approx 0$. Then, at the reached equilibrium of the activities, the weights are updated via common neural network optimisers such as stochastic GD or Adam (Eq. \ref{eq3}). This process is repeated for every training step, typically for a given data batch. Inference is typically performed by standard GD on the energy, which can be seen as the Euler discretisation of the gradient system $\dot{\mathbf{z}_\ell} = - \partial \mathcal{F} / \partial \mathbf{z}_\ell$. \textsc{JPC} simply leverages well-tested ODE solvers to integrate this gradient flow.
\end{tcolorbox}

\section{Runtime efficiency} \label{runtime}
A comprehensive benchmarking of various types of PCN with GD as inference optimiser was recently performed by \cite{pinchetti2024benchmarking}. For this reason, here we focus on runtime efficiency, comparing standard Euler integration of the inference gradient flow dynamics with Heun, a second-order explicit Runge–Kutta method. Note that, as a second-order method, Heun has a higher computational cost than Euler; however, it could still be faster if it requires significantly fewer steps to converge.

The solvers were compared on feedforward networks trained to classify standard image datasets, with different number of hidden layers $H \in \{3, 5, 10\}$. Because our goal was to specifically test for runtime, we trained each network for only one epoch across different step sizes $dt \in \{5e^{-1}, 1e^{-1}, 5e^{-2}\}$, retaining the run with the highest mean test accuracy achieved (see Figures \ref{fig3}-\ref{fig6}). Unlike Euler, Heun employed a standard Proportional–Integral–Derivative step size controller. Therefore, to make comparison fair, we also trained networks across many upper integration limits $T \in \{5, 10, 20, 50, 100, 200, 500\}$, again retaining the run with the maximum accuracy (Figures \ref{fig3}-\ref{fig6}). In cases where the accuracy difference between any $T$ was not significantly different, we selected runs with the smaller $T$.

\begin{figure*}[t]
    \begin{center}
        \centerline{\includegraphics[width=\textwidth]{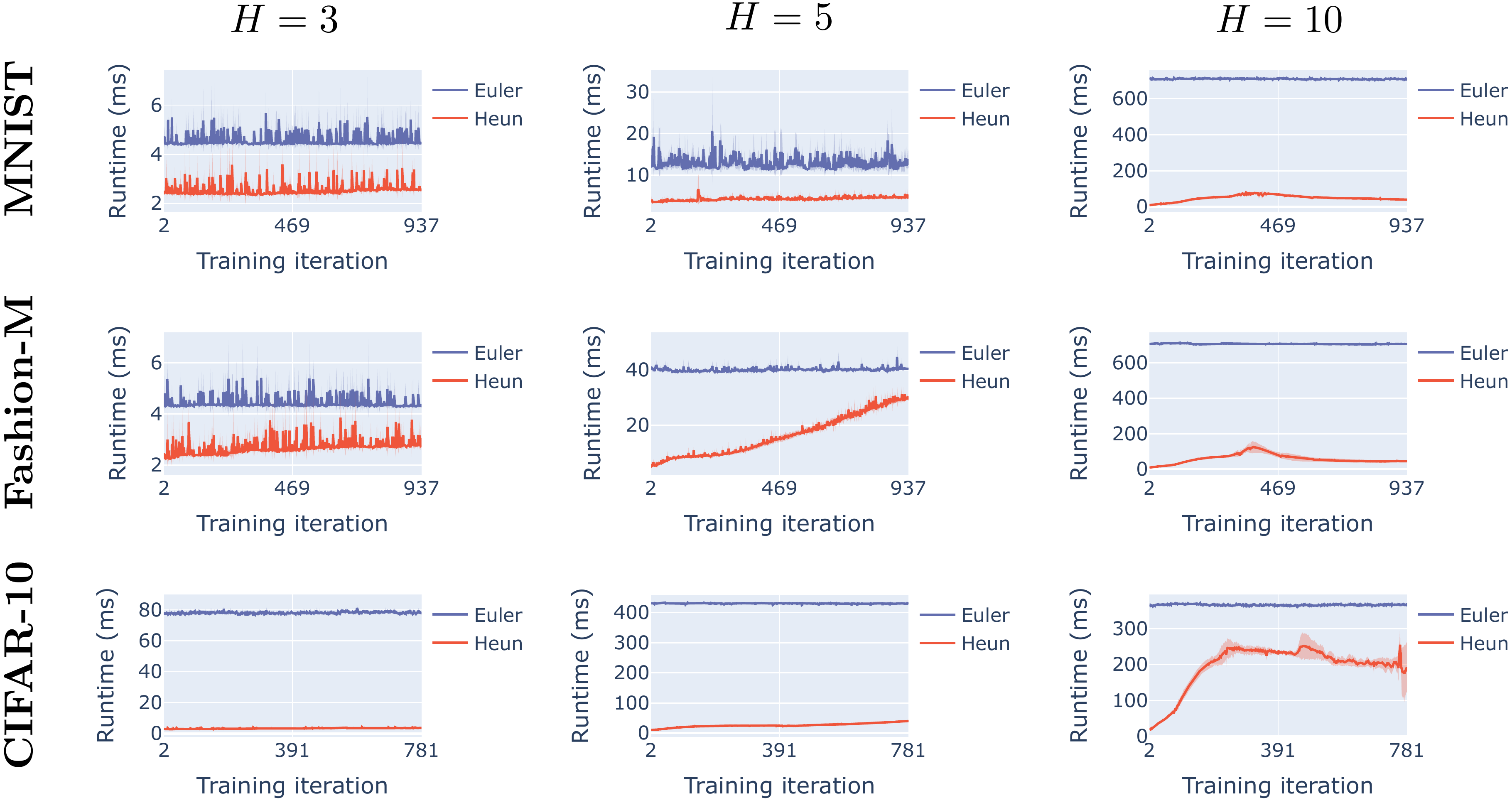}}
        \caption{\textbf{Second-order Runge–Kutta method (Heun) solves PC inference faster than standard Euler on a range of datasets and networks.} We plot the wall-clock time of Euler and Heun at each training step of one epoch for networks with hidden layers $H \in \{3, 5, 10\}$ trained on standard image classification datasets. The runs with the highest mean test accuracy achieved across different hyperparameters were selected (see Figures \ref{fig3}-\ref{fig6}). The time of the first training iteration where ``just-in-time'' (jit) compilation occurs is excluded. All networks had 300 hidden units and Tanh as activation function, and were trained with learning rate $1e^{-3}$ and batch size $64$. Shaded regions indicate $\pm1$ standard deviation across 3 different random weight initialisations.}
        \label{fig1}
    \end{center}
    \vskip -0.2in
\end{figure*}

Figure \ref{fig1} shows that, despite requiring more computations at each step, Heun tends to converge significantly faster than Euler, and in general more so on deeper networks ($H=10$). The convergence behaviour of Euler was more consistent during training across datasets and network depths, with Heun sometimes increasing in runtime. As a caveat, we also note that other optimiser-specific hyperparameters could lead to different results, and we welcome the community to test these and other solvers against other implementations as well as tasks.

\section{Implementation} \label{implement}
\textsc{JPC} provides both a simple, high-level application programming interface (API) to train and test PCNs in a few lines of code (\S\ref{basic-api}) and more advanced functions offering greater flexibility as well as additional features (\S\ref{advanced-api}). It is built on top of three main JAX libraries:
\begin{itemize}
    \item \href{https://github.com/patrick-kidger/equinox}{\texttt{Equinox}} \cite{kidger2021equinox}, to define neural networks with PyTorch-like syntax,
    \item \href{https://github.com/patrick-kidger/diffrax}{\texttt{Diffrax}} \cite{kidger2022neural}, to leverage ODE solvers to integrate the gradient flow PC inference dynamics (Eq. \ref{eq2}), and
    \item \href{https://github.com/google-deepmind/optax}{\texttt{Optax}} \cite{deepmind2020jax}, for parameter optimisation (Eq. \ref{eq3}).
\end{itemize}
Below we provide a sketch of \textsc{JPC} with pseudocode, referring the reader to the \href{https://thebuckleylab.github.io/jpc/}{documentation} and the \href{https://thebuckleylab.github.io/jpc/examples/discriminative_pc/}{example notebooks} for more details.

\subsection{Basic API}  \label{basic-api}
The high-level API allows one to update the parameters of essentially any \href{https://github.com/patrick-kidger/equinox}{\texttt{Equinox}} network with PC in a single function call \code{ jpc.make\_pc\_step }.

\begin{jupyterCell}
from (*@\textcolor{codeblue}{jpc}@*) import make_pc_step

result (*@\textcolor{codepurple}{=}@*) make_pc_step(
    model,        # an equinox model with callable layers
    optim,        # optax optimiser
    opt_state,    # optimiser state
    y,            # some target (observation)
    x             # optional input (prior)
)

# updated model and optimiser
model (*@\textcolor{codepurple}{=}@*) result["model"]
optim, opt_state (*@\textcolor{codepurple}{=}@*) result["optim"], result["opt_state"]
\end{jupyterCell}
As shown above, at a minimum \code{ jpc.make\_pc\_step } takes a model, an \href{https://github.com/google-deepmind/optax}{\texttt{Optax}} optimiser and its 
state, and some data. For a model to be compatible with PC updates, it needs to be split into callable layers (see the 
\href{https://thebuckleylab.github.io/jpc/examples/discriminative_pc/}{example notebooks}). Also note 
that an input is actually not needed for unsupervised training. In fact, 
\code{ jpc.make\_pc\_step } can be used for classification and generation tasks, for 
supervised as well as unsupervised training (again see the \href{https://thebuckleylab.github.io/jpc/examples/discriminative_pc/}{example notebooks}). 

Under the hood, \code{ jpc.make\_pc\_step }:
\begin{enumerate}
    \item integrates the gradient flow PC inference dynamics (Eq. \ref{eq2}) using a \href{https://github.com/patrick-kidger/diffrax}{\texttt{Diffrax}} ODE solver (Heun by default), and
    \item updates the parameters at the numerical solution of the activities (Eq. \ref{eq3}) with a given \href{https://github.com/google-deepmind/optax}{\texttt{Optax}} optimiser.
\end{enumerate}
Default parameters such as the ODE solver and a step size controller can all be changed. One has also the option of recording a variety of metrics including the energies and activities at each inference step (see the \href{https://thebuckleylab.github.io/jpc/}{documentation} for more details). 

Importantly, \code{ jpc.make\_pc\_step } is already ``jitted'' for performance, and the user only needs to embed this function in a data loop to train a neural network. We also provide convenience, already-jitted functions for testing specific PC models, including \code{ jpc.test\_discriminative\_pc } and \code{ jpc.test\_generative\_pc }.

A similar API is provided for hybrid PC (HPC) models \cite[see][]{tscshantz2023hybrid} with \code{ make\_hpc\_step }
\begin{jupyterCell}
from (*@\textcolor{codeblue}{jpc}@*) import make_hpc_step

result (*@\textcolor{codepurple}{=}@*) make_hpc_step(
    generator,          # generative model
    amortiser,          # model for inference amortisation
    optims,             # optimisers, one for each model
    opt_states,         # optimisers' state
    y,                 
    x 
)

\end{jupyterCell}
where now one has to pass an additional model (and associated optimiser objects) for amortising the inference of the generative model. Again, there is an option to override the default ODE solver parameters and record different metrics, and a convenience function for testing HPC \code{ jpc.test\_hpc } is also provided. We refer to the \href{https://thebuckleylab.github.io/jpc/examples/hybrid_pc/}{example notebook on HPC} for more details.

\subsection{Advanced API} \label{advanced-api}
While convenient and abstracting away many of the details, the basic API can be limiting, for example if one would like to jit-compile some additional computations within each PC training step. Advanced users can therefore access all the underlying functions of the basic API as well as additional features.

\paragraph{Custom step function.} A custom PC training step would look like the following.
\begin{jupyterCell}
import jpc

# 1. initialise activities with a feedforward pass
activities (*@\textcolor{codepurple}{=}@*) jpc.init_activities_with_ffwd(model, x)

# 2. run inference to equilibrium (Eq. (*@\ref{eq2}@*))
equilibrated_activities (*@\textcolor{codepurple}{=}@*) jpc.solve_inference(
    params(*@\textcolor{codepurple}{=}@*)(model, None),
    activities(*@\textcolor{codepurple}{=}@*)activities,
    output(*@\textcolor{codepurple}{=}@*)y,
    (*@\textcolor{black}{input}@*)(*@\textcolor{codepurple}{=}@*)x
)

# 3. update parameters at the activities' solution (Eq. (*@\ref{eq3}@*))
update_result (*@\textcolor{codepurple}{=}@*) jpc.update_params(
    params(*@\textcolor{codepurple}{=}@*)(model, None),
    activities(*@\textcolor{codepurple}{=}@*)equilibrated_activities,
    optim(*@\textcolor{codepurple}{=}@*)optim,
    opt_state(*@\textcolor{codepurple}{=}@*)opt_state,
    output(*@\textcolor{codepurple}{=}@*)y,
    (*@\textcolor{black}{input}@*)(*@\textcolor{codepurple}{=}@*)x
)
\end{jupyterCell}
This can be embedded in a jitted function with any other additional computations. One has also the option of using any \href{https://github.com/google-deepmind/optax}{\texttt{Optax}} optimiser to perform inference. In addition, the user can access other initialisation methods for the activities, the standard energy functions for PC and HPC, and the activity as well as parameter gradients used by the update functions. In fact, this is all there is to \textsc{JPC}, providing a simple framework to extend the library.
\begin{figure*}[h]
    \begin{center}
        \centerline{\includegraphics[width=\textwidth]{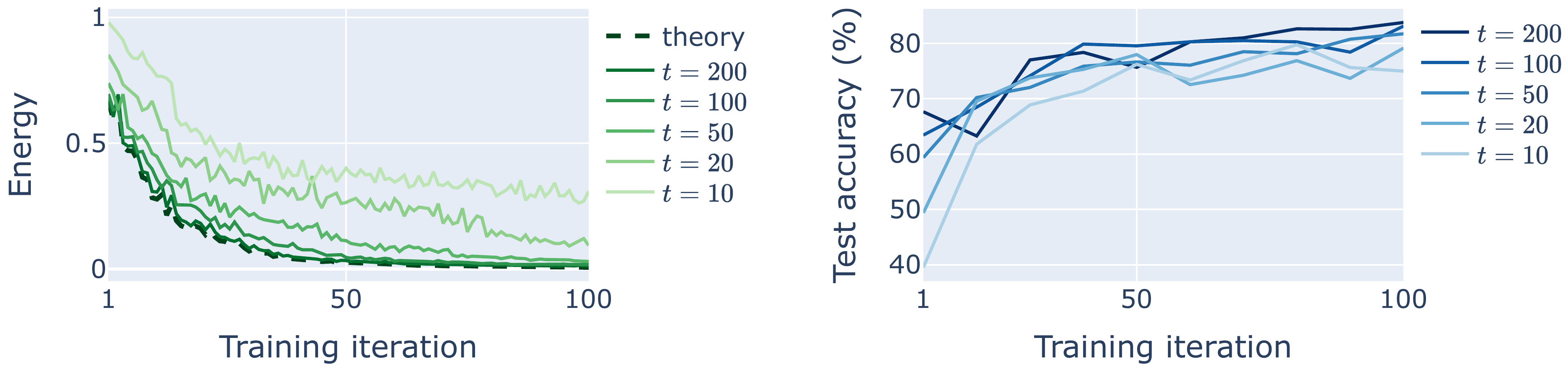}}
        \caption{\textbf{Theoretical PC energy for deep linear networks (Eq. \ref{eq4}) can help predict whether more inference could lead to better performance.} We compare the theoretical energy with the numerical energy for different upper limits $t$ of inference integration, as well as test accuracies, for a 10-hidden-layer, 300-width linear network trained to classify MINST with learning rate $1e^{-3}$ and batch size $64$. Results were consistent across different random initialisations.}
        \label{fig2}
    \end{center}
    \vskip -0.2in
\end{figure*}
\paragraph{Theoretical tools.} \textsc{JPC} also comes with some analytical tools that can be used to both study, and potentially diagnose issues with, PCNs. We now demonstrate this with an example. For deep linear networks (where $f_\ell$ is the identity at every layer), \cite{innocenti2024only} showed that the energy (Eq. \ref{eq1}) at the inference equilibrium $\nabla_{\mathbf{z}} \mathcal{F} = \mathbf{0}$ has the following closed-form solution as a rescaled mean squared error loss
\begin{equation}
    \mathcal{F}^* = \frac{1}{2N} \sum_{i=1}^N (\mathbf{y}_i - \matr{W}_{L:1}\mathbf{x}_i)^T \matr{S}^{-1}(\mathbf{y}_i - \matr{W}_{L:1}\mathbf{x}_i)
    \label{eq4}
\end{equation}
where the rescaling is $\matr{S} = \matr{I}_{d_y} + \sum_{\ell=2}^L (\matr{W}_{L:\ell})(\matr{W}_{L:\ell})^T$, and we use the shorthand $\matr{W}_{k:\ell} = \matr{W}_k \dots \matr{W}_\ell$ for $\ell, k \in 1,\dots, L$.

\cite{innocenti2024only} found Eq. \ref{eq4} to perfectly predict the energy (Eq. \ref{eq1}) at numerical convergence when $\nabla_{\mathbf{z}} \mathcal{F} \approx \mathbf{0}$. Figure \ref{fig2} suggests that the theoretical energy can also help determine whether sufficient inference has been performed, in that more inference steps seem to correlate with better test accuracy on MNIST. Similar results are observed on Fashion-MNIST (see Figure \ref{fig7}).

\section{Conclusion} \label{conclusion}
We introduced \textsc{JPC}, a new JAX library for training a variety of PCNs. Unlike existing frameworks, \textsc{JPC} is extremely simple (<1000 lines of code), completely functional in design, and leverages well-tested ODE solvers to integrate the gradient flow inference dynamics of PCNs. We showed that a second-order solver can provide significant speed-ups in runtime over standard Euler integration across a range of datasets and networks. As a straightforward extension, it would be interesting to integrate stochastic differential solvers, which recent work associates with better generation performance \cite{zahid2023sample, oliviers2024learning}. We hope that, together with other recent PC libraries \cite{pinchetti2024benchmarking, legrand2024pyhgf}, \textsc{JPC} will help facilitate research on PCNs.

\section*{Acknowledgements}
FI would like to thank Patrick Kidger for early advice on the use of \texttt{Diffrax} and acknowledges funding from the Sussex Neuroscience 4-year PhD Programme. PK is funded by the European Innovation Council (EIC) Pathfinder Challenges, Project METATOOL with Grant Agreement (ID: 101070940). WYF is funded by the European Research Council (ERC) Advanced Investigator Grant CONSCIOUS (ID: 101019254). MdLV was supported by VERSES AI. RS was supported by the Leverhulme Trust through the be.AI Doctoral Scholarship Programme in biomimetic embodied AI. CLB was partially supported by the European Innovation Council (EIC) Pathfinder Challenges, Project METATOOL with Grant Agreement (ID: 101070940).

\section*{Author contributions} \label{contributions}
FI wrote all the library code, ran the experiments, and wrote the paper. PK came up with the idea of using ODE solvers to integrate the gradient flow PC inference dynamics. WYF and MdLV helped test the library, and RS and CLB contributed to conceptual discussions.


\medskip

\bibliography{references}

\newpage
\appendix

\section{Supplementary figures} \label{supp-figs}
\begin{figure*}[h]
    \begin{center}
        \centerline{\includegraphics[width=\textwidth]{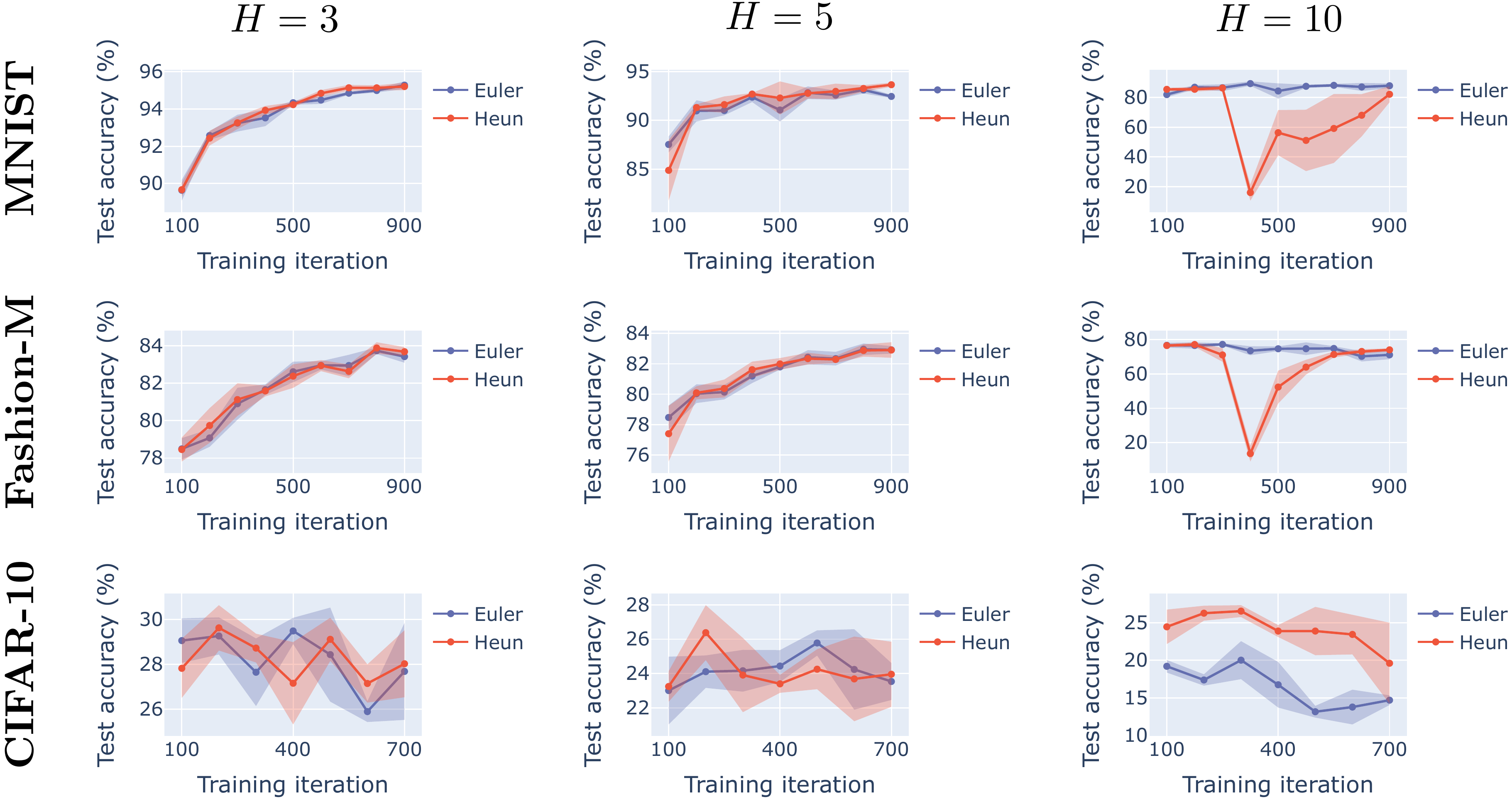}}
        \caption{\textbf{Test accuracies for Figure \ref{fig1}.} These accuracies were selected from Figures \ref{fig4}-\ref{fig6} based on the lowest upper integration limit $T$ at which the maximum mean accuracy was achieved. Note that the experiments were not optimised for accuracy, since we were specifically interested in the runtime of different ODE solvers at comparable performance. We refer to \cite{pinchetti2024benchmarking} for a comprehensive performance benchmarking of PCNs.}
        \label{fig3}
    \end{center}
\end{figure*}
\begin{figure*}[h]
    \begin{center}
        \centerline{\includegraphics[width=\textwidth]{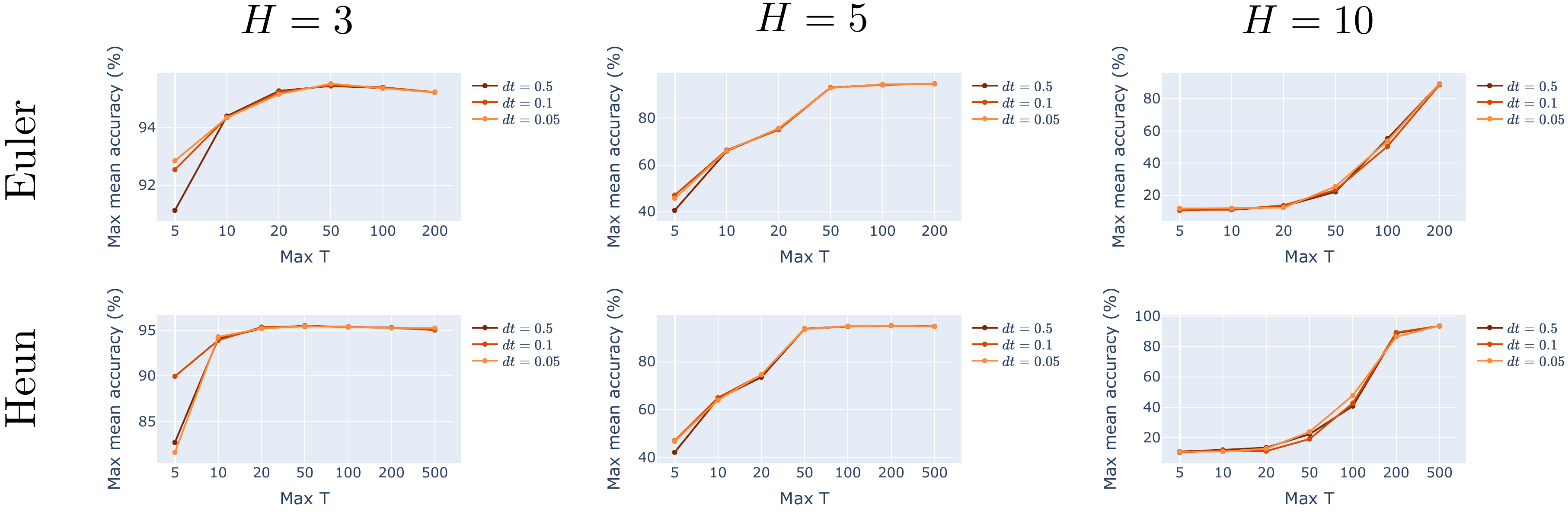}}
        \caption{\textbf{Maximum mean test accuracy on MNIST achieved with Euler and Heun as a function of different step sizes $dt$ and upper integration limits $T$.} For the results in Figure \ref{fig1} with $H=3$, we selected runs with $T=20$, and $dt=0.5$ for Euler and $dt=0.05$ for Heun. For $H=5$, we selected $T=50$, and $dt=0.5$ for Euler and $dt=0.05$ for Heun. Finally, for $H=10$, $T=200$ and $dt=0.05$ were chosen for both solvers.}
        \label{fig4}
    \end{center}
\end{figure*}
\begin{figure*}[h]
    \begin{center}
        \centerline{\includegraphics[width=\textwidth]{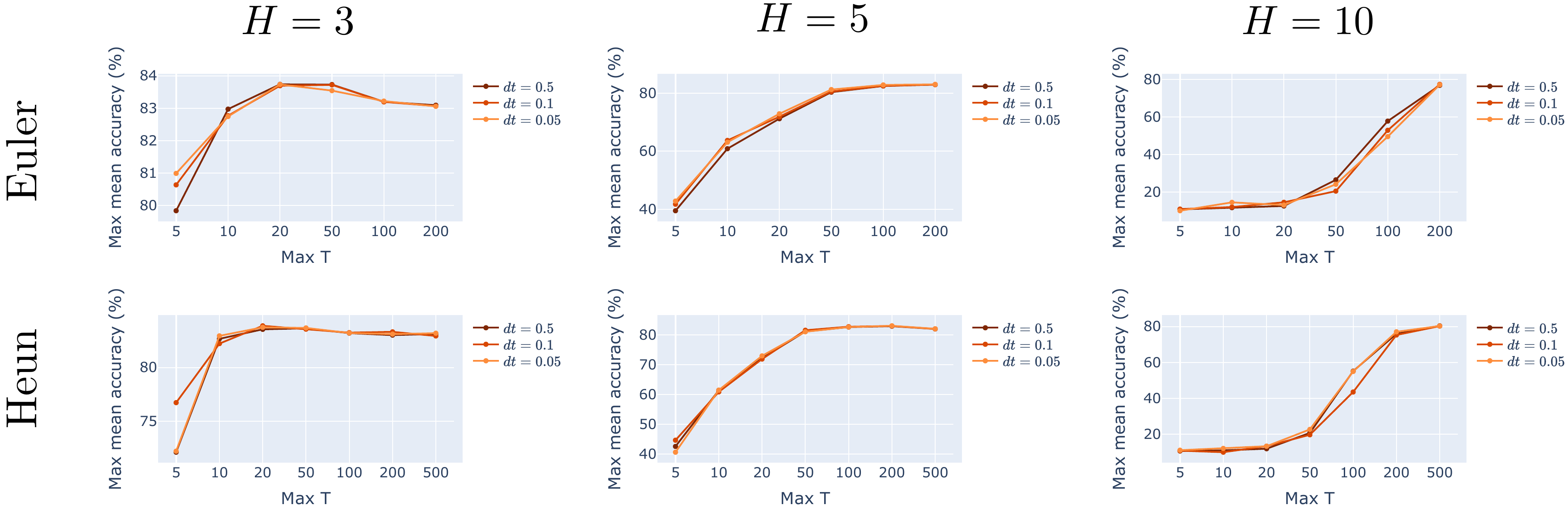}}
        \caption{\textbf{Same results as Figure \ref{fig4} for Fashion-MNIST.} For the results in Figure \ref{fig1} with $H=3$, we selected runs with $T=20$, and $dt=0.5$ for Euler and $dt=0.1$ for Heun. For the other network depths, the same hyperparameters were chosen for both solvers: $T=200$ and $dt=0.5$ for $H=5$, and $T=200$, and $dt=0.05$ for $H=10$.}
        \label{fig5}
    \end{center}
\end{figure*}
\begin{figure*}[t]
    \begin{center}
        \centerline{\includegraphics[width=\textwidth]{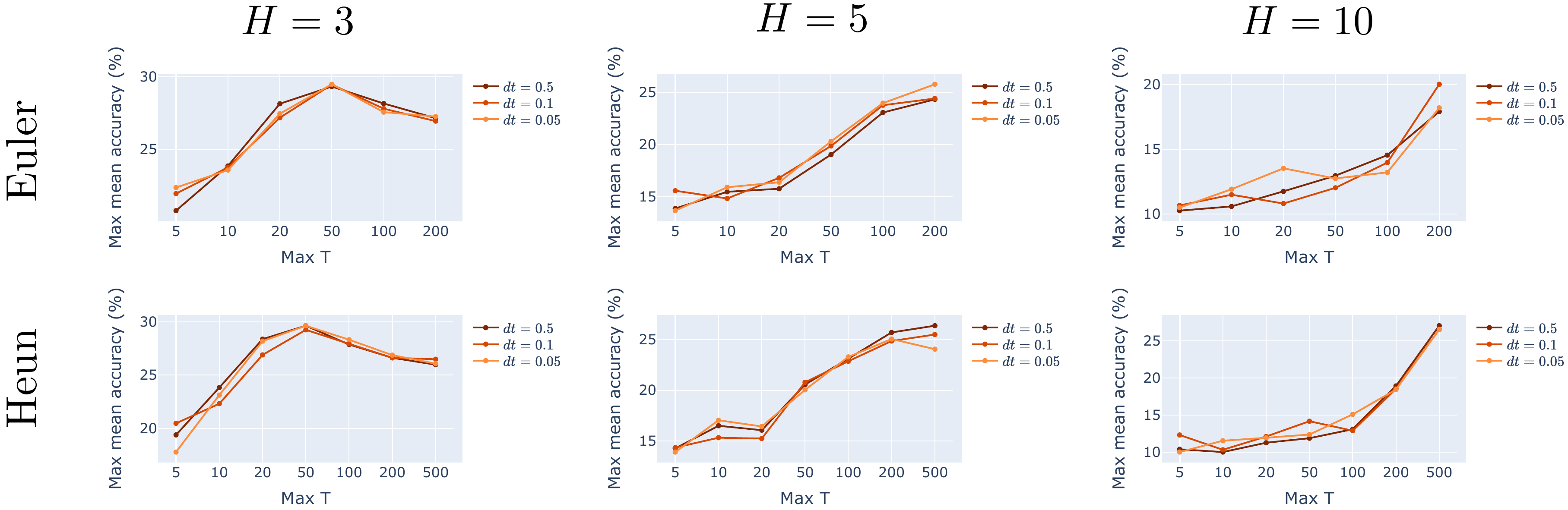}}
        \caption{\textbf{Same results as Figure \ref{fig4} for CIFAR-10.} For the results in Figure \ref{fig1} with $H=3$, we selected runs with $T=50$ and $dt=0.05$ for both solvers. For $H=5$, we selected $T=200$ and $dt = 0.05$ for Euler, and $T=500$ and $dt=0.5$ for Heun. Finally, for $H=10$, we selected $dt=0.1$, with $T=200$ for Euler and $T=500$ for Heun.}
        \label{fig6}
    \end{center}
\end{figure*}
\begin{figure*}[h]
    \begin{center}
        \centerline{\includegraphics[width=\textwidth]{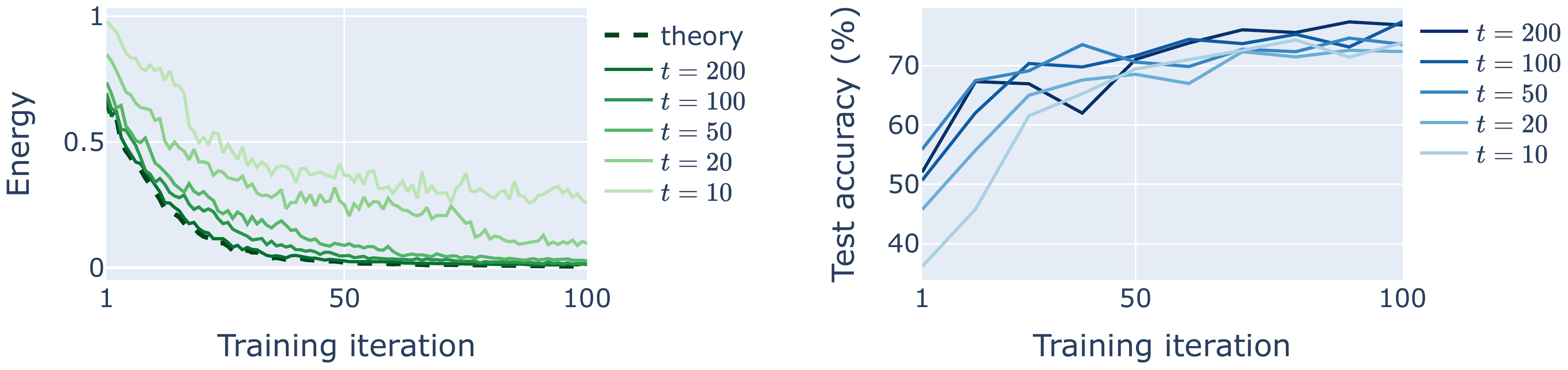}}
        \caption{\textbf{Same results as Figure \ref{fig2} for Fashion-MNIST.}}
        \label{fig7}
    \end{center}
\end{figure*}


\end{document}